# Orthogonal Policy Gradient and Autonomous Driving Application

Mincong Luo[1st*], Yin Tong[1st*] and Jiachi Liu[3rd*]
*Department of Information Centre*
*China Institute of Atomic Energy*
*Peking, China*
{luomincentos&ytongLisa&jcliu97}@ciae.ac.cn

*Abstract*—One less addressed issue of deep reinforcement learning is the lack of generalization capability based on new state and new target, for complex tasks, it is necessary to give the correct strategy and evaluate all possible actions for current state. Fortunately, deep reinforcement learning has enabled enormous progress in both subproblems: giving the correct strategy and evaluating all actions based on the state.

In this paper we present an approach called orthogonal policy gradient descent(OPGD) that can make agent learn the policy gradient based on the current state and the actions set, by which the agent can learn a policy network with generalization capability. we evaluate the proposed method on the 3D autonomous driving enviroment TORCS compared with the baseline model, detailed analyses of experimental results and proofs are also given.

## I. INTRODUCTION

In this paper we proposed a deep reinforcement learning(DRL) method called orthogonal policy gradient descent, which is prooved that the global optimization objective function can reach maximum value and is used in the application of autonomous driving.

We prooved that the average-reward $\rho$ meets $\frac{\partial \rho}{\partial \theta} = \sum_a d^\pi(s) \sum_a \frac{\partial \pi(a|s)}{\partial \theta} Q(a, s)$ in the **Theorem.1**, so the goal is to maximize the average-reward $\rho$ which means we can let $\left\|\frac{\partial \rho}{\partial \theta}\right\| = 0$ according to the principle of convex optimization. So the core idea of this paper is to make gradient vector $\frac{\partial \pi(a|s)}{\partial \theta}$ and $Q(\alpha,s)$ meet orthogonal(orthogonal policy gradient descent).

We propose a novel architecture with the combination of policy network $\pi(a|s)$ and Q-value network $Q(a, s)$ and learning strategy(orthogonal policy gradient descent) that leads to optimize the global objective function. We use this proposed learning algorithm for autonomous driving and test the feasibility and effectiveness of this algorithm in a 3D virtual environment which is called Torcs, in which the baseline method Actor-Critic algorithm based on DDPG(Deep Deterministic Policy Gradient) [10] is used to compared to our method.

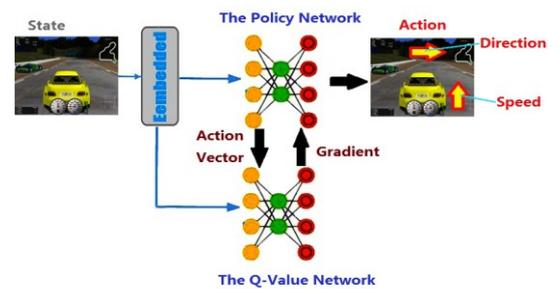

Figure 1. The framework of the proposed method to implement the autonomous driving.

## II. RELATED WORK

The related work includes the deep reinforcement learning, autonomous driving, and policy gradient optimization.

- Deep Reinforcement Learning(DRL): This approach extends reinforcement learning to the entire process from observation to action (sensors to motors or end to end) by forming it using a deep network and without explicitly designing state space or action space, in recent years DRL makes impressed success in automatic control and robotics[1][5][7].

- Autonomous Driving: Many different approaches for autonomous driving have been studied in recent years and the most ones are being engineered on deep learning methods[6][7][8], which often rely on virtual 3D driving environment for model training and simulation.

- Policy Gradient Optimization: Policy gradient methods are a notable section to the reinforcement learning tasks. Starting with the pioneering work in the early 1990s, these methods have been applied to a variety of robot learning problems ranging from simple control tasks to complex tasks like autonomous driving[4][1][9].

---

## III. PRELIMINARIES OF DEEP REINFORCEMENT LEARNING

In recent years the deep reinforcement learning(DRL) methods have shown promising results in virtual enviroment tasks and real-world tasks [13][12][8][9]. The study of reinforcement learning is based on the scope of classical physics. The time of the world can be divided into time slices, and there is a complete sequence, which is the series of time states mentioned above, one of the important assumptions is that each time the parameter adjustment has a deterministic effect on the environment, ie the input is determined and the output is determined[6][4][12].

$$\{s_1, a_1, r_1, s_2, a_2, r_2 ...\}$$

The Bellman equation[2] illustrates the relationship between the value function of the current state and the value function of the next state. From a formula perspective, the value of the current state is related to the value of the next step and the current feedback, Reward [1][2][5]. It shows that the Value Function can be calculated by iteration, which is basis for intensive learning.

$$v(s) = E[R_{t+1} + \lambda v(S_{t+1}) | S_t = s]$$

The Action-value function Q(a, s) is defined to denote the cumulative reward of strategy π starting from state s, after performing action a:

$$Q(a,s) = E[r_{t+1} + \lambda r_{t+2} + \lambda^2 r_{t+3} + ... | s, a] \quad (1)$$
$$= E_{s'}[r + \lambda Q^\pi(s', a') | s, a] \quad (2)$$

Q-Learning is based on value iteration. There is virtually no way to traverse all states, and all the actions, but only a limited series of samples. Therefore, only limited samples can be used for operation. A way to update the Q value is proposed in Q-Learning:

$$Q(S_t, A_t) \leftarrow Q(S_t, A_t) + \alpha(R_{t+1} + \lambda \max_a Q(S_{t+1}, a) - Q(S_t, A_t))$$

We define ρ(π) as the average reward formulation that is useful in the next, in which policies are ranked according to their long-term expected reward per step:

$$\rho(\pi) = \lim_{n \to \infty} \frac{1}{n} \{r_1 + r_2 + ... + r_n | \pi\} = \sum_s d^\pi(s) \sum_a \pi(a|s) R_s^a$$

Where the $d^\pi(s) = \lim_{t \to \infty} \Pr(s_t = s | s_0, \pi)$ is the stationary distribution of the all states under policy π.

## IV. THE THEOREM OF POLICY GRADIENT

The policy gradient $\frac{\partial \pi(a|s)}{\partial \theta}$ is what we wanna the agent to learn because the agent can update the policy based on the new states and targets by which. But first of all establishing the relationship between the policy gradient $\frac{\partial \pi(a|s)}{\partial \theta}$ and the globle value is needed. To address this issue the folllowing theorem is proposed.

*Theorem 1*: In the MDP(Markov Decision Process) for average-reward formulations:

$$\frac{\partial \rho}{\partial \theta} = \sum_a d^\pi(s) \sum_a \frac{\partial \pi(a|s)}{\partial \theta} Q(a,s)$$

*Proof 4.1:* For the average-reward formulations, we can denote the reward as:

$$V^\pi(s) = \sum_a \pi(a|s) Q(a,s)$$

so we can get:

$$\frac{\partial V^\pi(s)}{\partial \theta} = \frac{\partial}{\partial \theta} \sum_a \pi(a|s) Q(a,s)$$
$$= \sum_a (\frac{\partial \pi(a|s)}{\partial \theta} Q(a,s) + \pi(a|s) \frac{\partial Q(a,s)}{\partial \theta})$$
$$= \sum_a (\frac{\partial \pi(a|s)}{\partial \theta} Q(a,s) + \pi(a|s) \frac{\partial}{\partial \theta} (R_a^s - \rho(\pi) + \sum_{s'} P_{ss'}^a V^\pi(s')))$$
$$= \sum_a (\frac{\partial \pi(a|s)}{\partial \theta} Q(a,s) + \pi(a|s)(-\frac{\partial \rho}{\partial \theta} + \sum_{s'} P_{ss'}^a \frac{\partial V^\pi(s')}{\partial \theta}))$$

Which can be sorted out and summing by $d^\pi$ as:

$$\sum_s d^\pi(s) \frac{\partial \rho}{\partial \theta} = \sum_s d^\pi(s) \sum_a \frac{\partial \pi(a|s)}{\partial \theta} Q(a,s)$$
$$+ \sum_s d^\pi(s) \sum_a \pi(a|s) \sum_{s'} P_{ss'}^a \frac{\partial V^\pi(s')}{\partial \theta}$$
$$- \sum_s d^\pi(s) \frac{\partial V^\pi(s)}{\partial \theta}$$

Using the stationary of $d^\pi$ and we can get:

$$\sum_s d^\pi(s) \frac{\partial \rho}{\partial \theta} = \sum_s d^\pi(s) \sum_a \frac{\partial \pi(a|s)}{\partial \theta} Q(a,s)$$
$$+ \sum_{s' \in S} d^\pi(s') \frac{\partial V^\pi(s')}{\partial \theta} - \sum_{s \in S} d^\pi(s) \frac{\partial V^\pi(s)}{\partial \theta}$$
$$\Rightarrow \frac{\partial \rho}{\partial \theta} = \sum_a d^\pi(s) \sum_a \frac{\partial \pi(a|s)}{\partial \theta} Q(a,s)$$

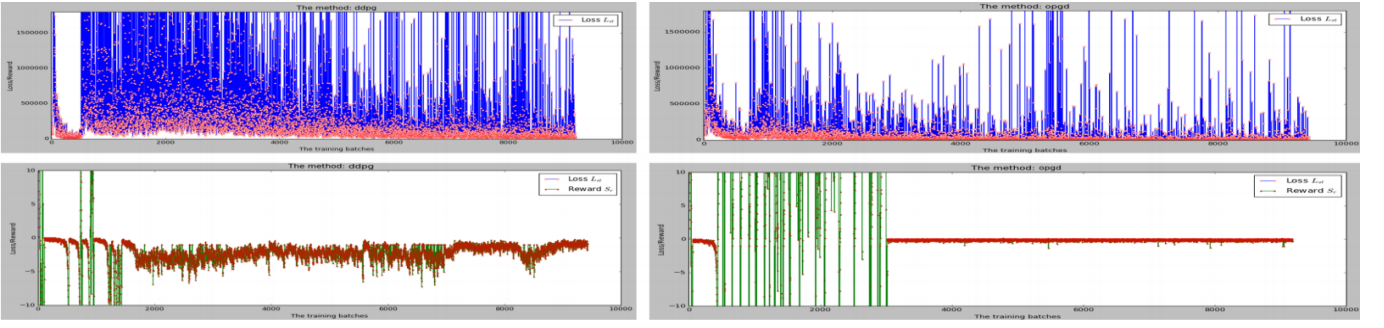

Figure 2. The loss curve and the reward curve of the proposed model(opgd) and the baseline model(ddpg).

Based on this theorem, we can use the $\frac{\partial \rho}{\partial \theta} = \sum_a d^\pi(s) \sum_a \frac{\partial \pi(a|s)}{\partial \theta} Q(a,s)$ to find the most suitable descent gradient. In the next section, we will propose a model for this.

## V. THE MODEL AND THE EXPERIMENT

---
**Algorithm 1** The training algorithm using orthogonal policy gradient with learning rate η

**Input:** The init QTable($s^{(0)}, a^{(0)}$); The init state $s^{(0)}$;
1: **for** t=1 to T do **do**
2:     $s^{(t)} \sim Env(s^{(t)}|a^{(t-1)})$;
3:     $a^{(t)} \sim \pi(a^{(t)}|s^{(t)}, \theta_a^{(t-1)})$;
4:     $QTable(s^{(t)}, a^{(t)}) \leftarrow Q(s^{(t)}, a^{(t)}, \theta_Q^{(t-1)})$;
5:     $L^{(t)} \leftarrow \sum_{s,a \in QTable} \frac{\partial \pi(a|s, \theta_a^{(t-1)})}{\partial \theta} Q(s, a, \theta_Q^{(t-1)})$;
6:     $\nabla_{\theta_Q}^{(t)} L^{(t)} \leftarrow \frac{\partial L^{(t)}}{\partial Q} \frac{\partial Q}{\partial \theta_Q}$;
7:     $\nabla_{\theta_a}^{(t)} L^{(t)} \leftarrow \frac{\partial L^{(t)}}{\partial Q} \frac{\partial Q}{\partial a} \frac{\partial \pi}{\partial \theta_a}$;
8:     $\theta_Q^{(t)} \leftarrow \theta_Q^{(t-1)} - \eta \nabla_{\theta_Q}^{(t)} L^{(t)}$;
9:     $\theta_a^{(t)} \leftarrow \theta_a^{(t-1)} - \eta \nabla_{\theta_a}^{(t)} L^{(t)}$;
10: **end for**
---

The $\rho(\pi)$ is defined as the average reward formulation denotes the performance of the polocies the agent chooses, which is an optimization target needs to be maximized. So the base idea is:

$$\frac{\partial \pi(a|s)^*}{\partial \theta} = \min_{\frac{\partial \pi(a|s)}{\partial \theta}} \rho(\pi)$$

Where the $\frac{\partial \pi(a|s)^*}{\partial \theta}$ means the gradient maximize the $\rho(\pi)$, so the core idea is let $\frac{\partial \rho}{\partial \theta} = 0$ and this means $\frac{\partial \pi(a|s)}{\partial \theta}$ and Q(a,s) need to be orthogonal:

**The Model:** The model with two approximation functions, $f^\nabla(\cdot)$ and $f^Q(\cdot)$, is proposed. The $f^\nabla(\cdot)$ tries to learn the map from the states and actions to the gradient, and $f^Q(\cdot)$ tries to evaluate the value of the actions based on the states. The whole structure of the model is given in the Figure 1.

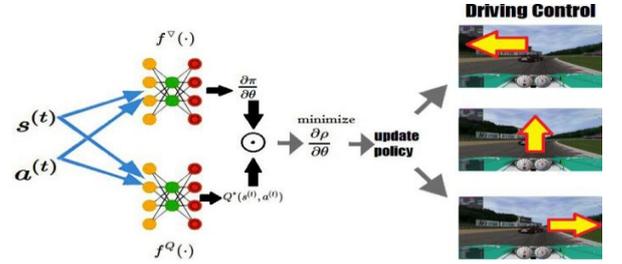

Figure 3. The model gives the detail actions(steering,acceleration,brake) based on the action a and state s with two approximation functions $f^\nabla(\cdot)$ and $f^Q(\cdot)$.

**The Experiment Enviroment:** We make the experiment enviroment on the Torcs(The Open Racing Car Simulator), which is an open source 3D car racing simulator available for Linux. For its openness, modularity and extensibility, TORCS has been adopted as a base for many research projects. examples include automated computation of car setups, human-assisted algorithmic generation of tracks and the application of several computing techniques (e.g. genetic programming) to different aspects of robot driving.

**The Baseline Model:** The baseline model we use is The Actor-Critic algorithm [5]. The Actor-Critic algorithm is essentially a hybrid method to combine the policy gradient method and the value function method together. The policy function is known as the actor, while the value function is referred to as the critic. Essentially, the actor produces the action a given the current state of the environment s, while the critic produces a signal to criticizes the actions made by the actor. We use the continuous Q-learning(SARSA)[11] as the critic model and use policy gradient method as the actor model.

**The Result:** The result shows that our method can better converge more stably and quickly compared to the baseline model. Compared with the baseline model, our model enables the loss function to converge better to a lower value interval and the fluctuation of loss becomes smaller. The change in reward value shows that the reward value of our model tends to be more stable and maintains a higher value range than the baseline model. In summary, our model performs better in this task.

## VI. CONCLUSION

In this work we developed a simple and effective model for deep reinforcement learning based on orthogonal policy gradient descent(OPGD). We demonstrated that the model has the generalization capability based on new state and new target. As experiments established, the approach is powerful and did better than specifically fitted solutions such as the weak-supervised of half-supervised models like Actor-Critic algorithm[2] in the application of autonomous driving. $f^{\nabla}(\cdot)$ is used to learn the policy gradient, which can be pre-trained as a network and used in the agent learning tasks. However, doubtlessly much more work can be done on this front by more tasks.

## VII. ACKNOWLEDGMENT

Thanks for the help from respected supervisor Jianyou Lv. $1^{st}*$ represent two jointed authors who contributed equally, thanks for the third author Jiachi Liu who belong to

## IX. APPENDIX

The bot controls the car in the game through a rather typical set of actuators, i.e., the steering wheel, the gas pedal, the brake pedal, and the gearbox. In addition, a meta-action is available to request a race restart to the server. The above table details the actions available and their representation.

TABLE I. DESCRIPTION OF THE AVAILABLE SENSORS. RANGES ARE REPORTED WITH THEIR UNIT OF MEASURE(WHERE DEFINED)

| Name | Range (unit) | Description |
|---|---|---|
| angle | $[-\pi,+\pi]$ (rad) | Angle between the car direction and the direction of the track axis. |
| curLapTime | $[0,+\infty]$ (s) | Time elapsed during current lap. |
| damage | $[0,+\infty]$ (point) | Current damage of the car (the higher is the value the higher is the damage). |
| distFromStart | $[0,+\infty]$ (m) | Distance of the car from the start line along the track line. |
| distRaced | $[0,+\infty]$ (m) | Distance covered by the car from the beginning of the race |
| focus | $[0,200]$ (m) | Vector of 5 range finder sensors. |
| fuel | $[0,+\infty]$ (l) | Current fuel level. |
| gear | $\{-1,0,1,\cdots 6\}$ | Current gear: -1 is reverse, 0 is neutral and the gear from 1 to 6. |
| lastLapTime | $[0,+\infty]$ (s) | Time to complete the last lap |
| opponents | $[0,200]$ (m) | Vector of 36 opponent sensors. |
| racePos | $\{1,2,\cdots,N\}$ | Position in the race with respect to other cars. |
| rpm | $[0,+\infty]$ (rpm) | Number of rotation per minute of the car engine. |
| speedX | $(-\infty,+\infty)$ (km/h) | Speed of the car along the longitudinal axis of the car. |
| speedY | $(-\infty,+\infty)$ (km/h) | Speed of the car along the transverse axis of the car. |
| speedZ | $(-\infty,+\infty)$ (km/h) | Speed of the car along the Z axis of the car. |
| track | $[0,200]$ (m) | Vector of 19 range finder sensors. |
| trackPos | $(-\infty,+\infty)$ | Distance between the car and the track axis. |
| wheelSpinVel | $[0,+\infty]$ (rad/s) | Vector of 4 sensors representing the rotation speed of wheels. |
| z | $[-\infty,+\infty]$ (m) | Distance of the car mass center from the surface of the track along the Z axis. |

TABLE II. DESCRIPTION OF THE AVAILABLE EFFECTORS.

| Name | Range | Description |
|---|---|---|
| accel | $[0,1]$ | Virtual gas pedal (0 means no gas, 1 full gas). |
| brake | $[0,1]$ | Virtual brake pedal (0 means no brake, 1 full brake). |
| clutch | $[0,1]$ | Virtual clutch pedal (0 means no clutch, 1 full clutch). |
| gear | -1,0,1,$\cdots$,6 | Gear value. |
| steering | $[-1,1]$ | Steering value: -1 and +1 means respectively full right and left, that corresponds to an angle of 0.366519 rad. |
| focus | $[-90,90]$ | Focus direction (see the focus sensors in degrees. |
| meta | 0,1 | This is meta-control command: 0 do nothing, 1 ask competition server to restart the race. |